\documentclass{jdmdh}
\usepackage{array}
\usepackage{pgfplots}
\usepackage{float}
\titlespacing*{\section}
{0pt}{4.5ex plus 1ex minus .2ex}{2.5ex plus .2ex}
\titlespacing*{\subsection}
{0pt}{4.5ex plus 1ex minus .2ex}{2.5ex plus .2ex}
\titlespacing*{\subsubsection}
{0pt}{4.5ex plus 1ex minus .2ex}{2.5ex plus .2ex}
\title{Rerunning OCR: A Machine Learning Approach to Quality Assessment and Enhancement Prediction}
\author[1]{Pit Schneider}
\author[1]{Yves Maurer}
\affil[1]{National Library of Luxembourg, Luxembourg} 

\corrauthor{Pit Schneider}{pit.schneider@bnl.etat.lu}

\begin{document}

\maketitle

\abstract{Iterating with new and improved OCR solutions enforces decision making when it comes to targeting the right candidates for reprocessing. This especially applies when the underlying data collection is of considerable size and rather diverse in terms of fonts, languages, periods of publication and consequently OCR quality. This article captures the efforts of the National Library of Luxembourg to support those targeting decisions. They are crucial in order to guarantee low computational overhead and reduced quality degradation risks, combined with a more quantifiable OCR improvement. In particular, this work explains the methodology of the library with respect to text block level quality assessment. Through extension of this technique, a regression model, that is able to take into account the enhancement potential of a new OCR engine, is also presented. They both mark promising approaches, especially for cultural institutions dealing with historical data of lower quality.}

\keywords{optical character recognition; quality assessment; enhancement prediction; candidate selection; machine learning; historical data; cultural institutions}

\section{Context}
In the context of its digitization program, the National Library of Luxembourg (BnL) started a first initiative in the optical character recognition (OCR) space in 2006. Back then the external scanning suppliers were in charge of performing OCR on the scanned historical newspapers, using various software solutions over the years. Although OCR is considered a largely solved problem for modern documents (\citet{doermann}), it remains a non-trivial task for historical data. That is why the library always considered the resulting output to feature a quality standard that could be improved in the future, with means of continuing software advancements.
\\ \\
A BnL pilot project conducted by \citet{maurer} proposed a framework to rerun OCR using a contemporary engine, such as Tesseract (\citet{kay}). The method leverages a metric that compares the new and original output on the ratio of number of characters belonging to words found in a dictionary. Altogether, the related article described promising results, served as a proof of concept and marked the starting point for subsequent OCR initiatives.
\\ \\
Fast forwarding to the year 2020, a new project is initiated, aiming to build a new in-house OCR model, referred to as \textsc{NewModel} in the rest of this article. It was trained on BnL data and represents an improvement on the original OCR quality. A prerequisite for the application of \textsc{NewModel}, however, is a method that is able to first assess the original OCR quality, without relying on any ground truth counterparts. In terms of terminology, this technique is referred to as \textit{automatic} OCR quality assessment. The motivation for employing such an approach and making it a prerequisite is threefold. It enables: \\
\begin{enumerate}
    \item The reduction of computation time through selective targeting of reprocessing candidates.
    \item The collection of statistical insights, estimating the improvement in OCR accuracy.
    \item The lowering of the risk of a potential accuracy reduction for a subset of the data.
\end{enumerate}
\begin{align}\label{enum:goals}
\end{align}
\subsection{Related Work} \label{subsection:relatedwork}
Although the motivations in (\ref{enum:goals}) seem compelling, tackling the problem of automatic quality assessment has seen relatively limited attention. More research has been devoted to the field of OCR post-correction techniques (e.g. \citet{neudecker}). Additionally, there are a number of studies (e.g. \citet{strien} and \citet{hill}) that aim to assess the impact of sub-optimal OCR quality on natural language processing (NLP) tasks, without necessarily measuring the quality itself. The general ideas for automatic quality assessment are diverse and do not seem to yield a clear winner method in terms of potential. Altogether, they can be split into three classes, depending on the used data source.
\\ \\
First, image based methods have probably seen the widest amount of research, but also turn out to be the most computationally expensive. By neglecting the possibly already existing OCR output and by exclusively looking at the source material, they try to analyze features that could impact OCR engines. For instance, \citet{blando} inspect specific typeface properties, while \citet{lu} aim to discover physical image distortions. Other examples include quantifying image degradation (\citet{peng}) and estimating the amount of blur (\citet{kieu}). Finally, a recent work from \citet{singh} uses surrogate models to learn document quality based on ground truth images.
\\ \\
Next, a common OCR engine produces output that extends beyond the raw text, which is the source for another class of approaches. In particular, \citet{gupta} label erroneous text bounding boxes based on their spatial distribution and geometry, as an indicator for OCR quality. Similarly, \citet{springmann} use engine confidence scores to compare the recognition quality of different models.
\\ \\
This leaves the processing of the output text only, which is the main source of inspiration for the present work. Here, the involvement of dictionaries has been looked at by researchers. For instance, \citet{alex} compare output words to the most similar entry in a dictionary. A very different approach comes from \citet{cavnar}, showing that n-grams can be successfully used to categorize texts by comparing them to n-gram based classification profiles. Moreover, work from \citet{kulp} and \citet{taghva} show the development of rule-base \textit{garbage token} detection systems. Finally, an original contribution from \citet{salah} uses a secondary (reference) OCR engine to cross-align output results using a Support Vector Regression technique.
\subsection{Data}
\begin{figure}[H]
\centering
\resizebox{0.53\textwidth}{!}{
    \begin{tikzpicture}
    \definecolor{c1}{RGB}{0,0,0}
    \begin{axis} [
        /pgf/number format/.cd,
        use comma,
        1000 sep={},
        xlabel={year},
        ylabel={number of newspaper issues per year},
        xmin=1841, xmax=1959,
        ymin=0, ymax=2000,
        xtick={1850, 1875, 1900, 1925, 1950},
        ytick={500, 1000, 1500},
        ymajorgrids=true,
        grid style=dashed,
    ]
    \addplot[
        color = c1,
        line width = 0.75mm,
        mark = none,]
        coordinates {
        (1841,51)(1842,49)(1843,51)(1844,181)(1845,291)(1846,153)(1847,156)(1848,413)(1849,440)(1850,363)(1851,368)(1852,364)(1853,365)(1854,362)(1855,359)(1856,435)(1857,494)(1858,722)(1859,635)(1860,592)(1861,902)(1862,929)(1863,1021)(1864,1080)(1865,1076)(1866,1068)(1867,1071)(1868,860)(1869,1098)(1870,1192)(1871,936)(1872,824)(1873,813)(1874,832)(1875,830)(1876,825)(1877,673)(1878,836)(1879,705)(1880,727)(1881,739)(1882,817)(1883,758)(1884,772)(1885,724)(1886,730)(1887,783)(1888,774)(1889,777)(1890,734)(1891,738)(1892,746)(1893,850)(1894,778)(1895,903)(1896,881)(1897,811)(1898,838)(1899,958)(1900,988)(1901,981)(1902,987)(1903,1031)(1904,1076)(1905,1134)(1906,1115)(1907,1203)(1908,1110)(1909,1195)(1910,1068)(1911,1079)(1912,1082)(1913,1225)(1914,1113)(1915,942)(1916,893)(1917,982)(1918,1072)(1919,1462)(1920,1328)(1921,1245)(1922,1255)(1923,1364)(1924,1702)(1925,1702)(1926,1548)(1927,1425)(1928,1380)(1929,1428)(1930,1397)(1931,1389)(1932,1314)(1933,1384)(1934,1393)(1935,1218)(1936,1237)(1937,1257)(1938,1288)(1939,1286)(1940,843)(1941,917)(1942,616)(1943,621)(1944,609)(1945,980)(1946,1227)(1947,1231)(1948,787)(1949,630)(1950,621)(1951,4)(1952,1)(1953,1)(1954,54)(1955,52)(1956,52)(1957,52)(1958,52)(1959,53)};
    \end{axis}
    \end{tikzpicture}
}
\caption{Publication date distribution of the BnL historical newspaper collection.}
\label{fig:yeardates}
\end{figure}
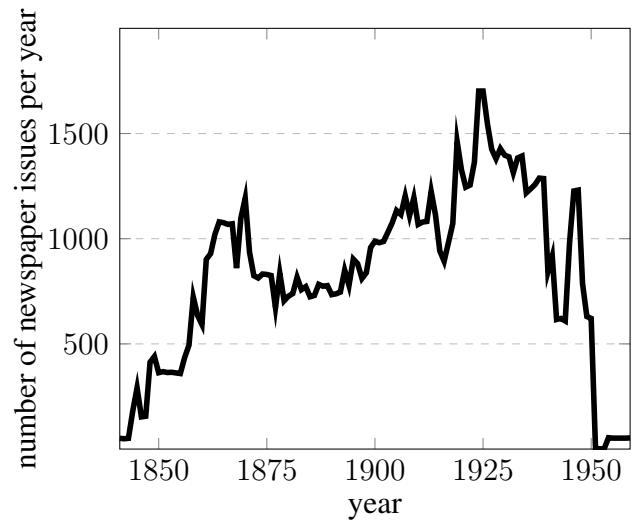

Subject to the application of \textsc{NewModel} are the approximately 102,000 historical newspaper issues, dating from 1841 to 1954. The newspaper articles are mostly written in German ($de$), French ($fr$) and Luxembourgish ($lb$). Their typography is more or less evenly split between Antiqua and Fraktur typefaces, rendering the data rather diverse (Figure~\ref{fig:yeardates}).
\\
\noindent This work frequently refers to \textit{blocks} as the most common data concept. A block represents the OCR output text, derived from the image of an individual paragraph or even a small article. As far as the layout is concerned, a block is always contained within a single text column. The choice to treat the data as a set of individual blocks is mainly motivated by the fact that there is a higher likelihood that properties, such as language or font, remain constant within a block.
\\ \\
The new OCR project has been initialized by first building a ground truth set. A subset of close to 7,000 text block images was selected and transcribed, mainly to serve for OCR training purposes. A second use case is to split-off of a separate test set, providing a foundation for automatic quality assessment. The possibility to test a given OCR output, by comparing it to its gold standard counterpart, is the basis for a supervised learning process. The resulting model will then be used for automatic quality assessment. Hence, finding a correlation between textual features, that can be computed without availability of a gold standard, and the text quality itself, is the venture discussed in the rest of this article.
\\ \\
From here on, this article will proceed by covering the two main contributions in the form of Sections \ref{section:qualityclassifier} and \ref{section:enhance}, before concluding by reflecting back on the statements in (\ref{enum:goals}). 
\newpage
\section{Quality Classifier} \label{section:qualityclassifier}
This work proposes a machine learning based classifier that is designed to assess the text quality of an entire text block. With this intention, a couple of definitions need to be established first.
\subsection{Definitions}
\begin{figure}[H]
\centering
\includegraphics[width=0.65\textwidth]{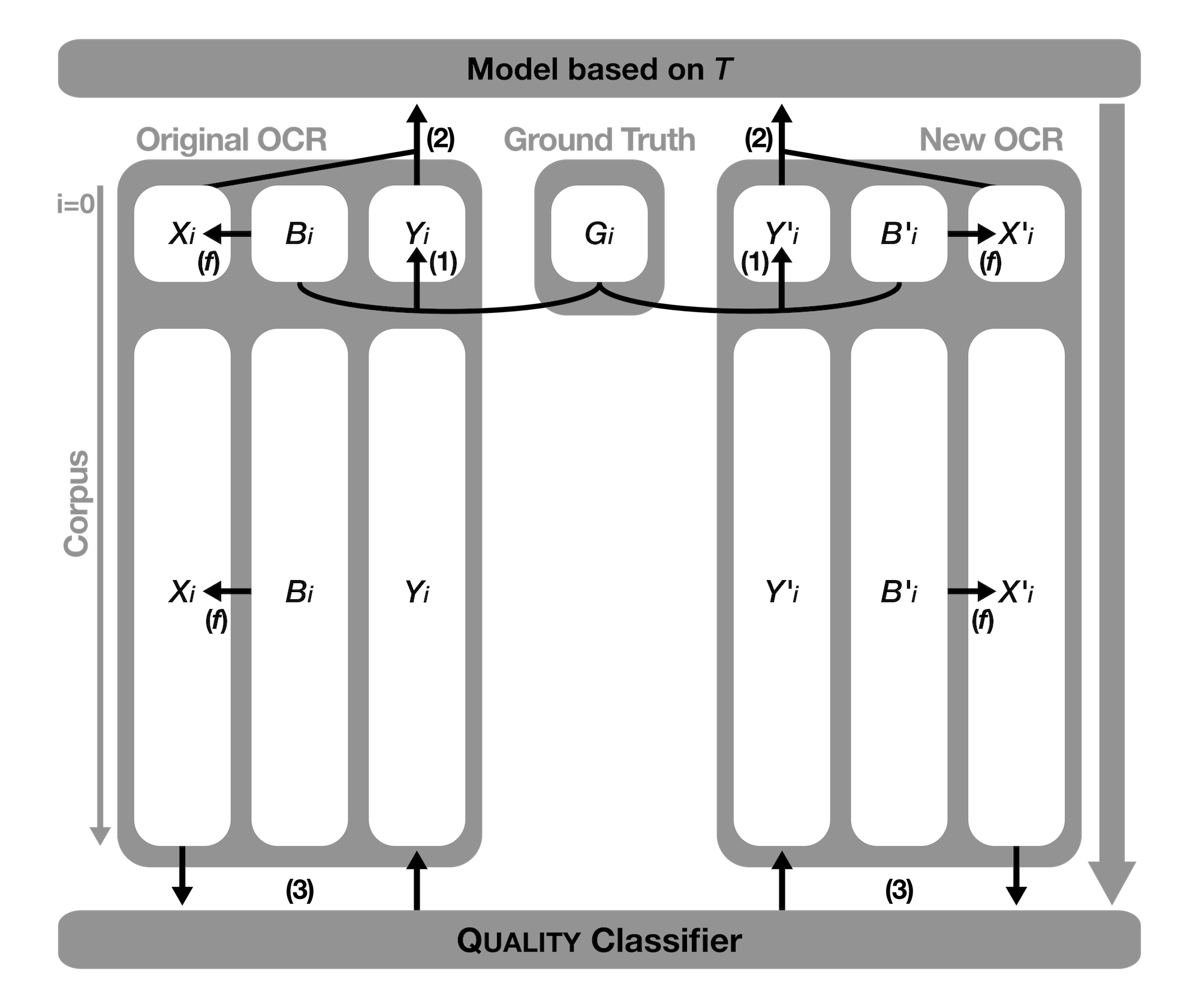}
\captionsetup{format=hang, margin=0.0\textwidth}
\caption{High-level workflow of ($f$) extracting text features, (1) determining the text quality, (2) training a model and (3) using that model to apply the classifier to the rest of the corpus.}
\label{fig:concept}
\end{figure}

Given the three motivations in (\ref{enum:goals}), fitting a binary classifier with classes referring to \textit{sufficient} and \textit{insufficient} quality is the logical starting point. That is why the classes space $C$ is defined as
\begin{align}
C = \{0, 1\},
\end{align}
with zero and one respectively referring to sufficient and insufficient quality. Coupling the positive class with bad OCR quality follows the notion of the classifier determining (a minority of) candidate blocks. \\ \\
A supervised learning process is using training data $T$, given by
\begin{align}
    T = \{(X_1, Y_1\!\in\!C), ..., (X_{n}, Y_{n}\!\in\!C)\}.
\end{align}
It is also defined that every feature vector\footnote{Please note that in this article exponents are always used as labels and should not be interpreted as mathematical powers.} has $k$ dimensions, such that
\begin{align}
    X_i = (x^0_i, x^1_i, ..., x^{k-1}_i).
\end{align}
The process of extracting all $k$ features from $i$th text block $B_i$ is referred to as the feature function
\begin{align}
    f: B_i \rightarrow X_i.
\end{align}
Having just defined $B_i$, there are a few additional notations related to a given block. While $G_i$ is used to refer to the ground truth version of $B_i$, the cardinality $|B_i|$ returns the total number of characters (including whitespaces) within the block. Furthermore, $B_i^t$ encodes all tokens (simple whitespace character delimitation) found in $B_i$. The concept of cardinality can again be utilized to obtain the length of a token. Lastly, the language function $\ell(B_i)$ returns the natural language of $B_i$.  \\ \\
The quality classifier can now be summarized as the function
\begin{align}
    \textsc{Quality}: B_i \rightarrow Y_i\!\in\!C.
\end{align}
To make \textsc{Quality} more robust it is trained on both the original OCR and \textsc{NewModel}
outputs, thus involving a variety of OCR software. This is illustrated in Figure~\ref{fig:concept}, which also shows the high-level workflow and some of the just established notations.
\subsection{Features} \label{subsection:features}
Next, focus is shifted to the topic of feature extraction. A collection of approaches and features has already been presented in Subsection~\ref{subsection:relatedwork}. While sampling them for \textsc{Quality}, BnL considered features that focus on revealing shortcomings in the original OCR engine (e.g. Antiqua/Fraktur typeface confusion), and not necessarily in the source image quality. A second consideration aspect was computational efficiency, preventing significant impacts on the processing time of the new OCR pipeline.
That is why the set retained for \textsc{Quality} is given by \\
\begin{itemize}
    \item $x^0$ : dictionary mapping,
    \item $x^1$ : tri-gram comparison,
    \item $x^2$ : garbage token detection,
    \item $x^3$ : publication year consideration. \\
\end{itemize}
Thus, $f$ operates on OCR output only. No feature is extracted from the source image, guaranteeing the efficiency of \textsc{Quality}.
\subsubsection{Dictionary Mapping}
A commonly used technique in automatic quality assessment is to compare the output words to a dictionary of the same language (e.g. \citet{alex}). Given a block $B_i$, its language $\ell(B_i)$, token $t\!\in\!B_i^t$ and dictionary $D^{\ell(B_i)}$, a binary variable is defined as \\
\begin{align}
    &map(t, D^{\ell(B_i)}) = 0\text{ if $t$ is not in the dictionary,} \nonumber \\
    &map(t, D^{\ell(B_i)}) = 1\text{ if $t$ is in the dictionary.}
\end{align}
In the context of \textsc{Quality}, the feature $x_i^0$ is derived from $B_i$ by computing ratio
\begin{align}
    x_i^0 = f(B_i)[0] = \frac{\sum_{t\in B_i^t}map(t, D^{\ell(B_i)})\times |t| }{\sum_{t\in B_i^t}|t|}. \label{formula:dict}
\end{align}
Given (\ref{formula:dict}), every token is weighted by its own length, instead of simply returning the fraction of successfully matched tokens.
\subsubsection{Tri-Gram Comparison}
As suggested by \citet{zipf}, given a natural language, the word rank-frequency distribution is a relation that is inversely proportional. The same law naturally holds for smaller entities within a language, such as n-grams. Building on this, \citet{cavnar} have demonstrated that texts can be categorized by comparing their contained n-grams to n-gram based classification profiles. \\ \\
In a similar way, an n-gram similarity measure is established for \textsc{Quality}. More specifically, the measure makes use of the ranks of the top $\gamma$ tri-grams in terms of frequency of language $\ell(B_i)$. The rank function $r(tri, \ell(B_i))$ returns the frequency rank of any tri-gram $tri$ for language $\ell(B_i)$. Before computing the feature value, all possible character tri-grams are extracted from every $t\!\in\!B_i^t$. It should be noted that tri-grams are limited to only span across letter characters. For instance, there is
\begin{align}
    t\!\in\!B_i^t &= Luxemb0urg \nonumber \\
    \text{tri-grams for }t &: \{lux, uxe, xem, emb, urg\}.
\end{align}
\\
Let $B^{tri}_i$ denote the set of all tri-grams in $B_i$. The feature value $x^1_i$ is calculated by
\begin{align} \label{formula:x2}
    x^1_i = f(B_i)[1] = 1\!-\!\frac{\sum_{tri\in B_i^{tri}}min\big(\gamma, r(tri, \ell(B_i))\big)}{\gamma\times|B_i^{tri}|}.
\end{align}
As a result of the exponential nature of the Zipfian distribution, the value of $\gamma$ seems rather inconsequential as long as it is not too small. During the implementation process, $\gamma\!=\!1000$ was chosen by BnL, safely covering all major tri-grams (in terms of importance) in the language. Naturally, the potential of this feature is increasing as $|B_i|$ increases as well.
\subsubsection{Garbage Token Detection}
As stated by \citet{wudtke}, a more serious category of OCR quality issues is the presence of tokens which also render the prediction of their correct replacement tokens infeasible. A feature, describing the amount of $garbage$ $tokens$ within $B_i$, combines ideas by \citet{kulp} and \citet{taghva} into a set of nine precise rules. \\ \\
A token $t\!\in\!B_i^t$ is identified as garbage in case it holds that $t$ contains at least one of the following: \\
\begin{enumerate}
    \item twenty-one characters.
    \item three consecutive occurrences of the same character.
    \item four consecutive vowels.
    \item six consecutive consonants.
    \item one vowel and at least one consonant and the count of one of them is more than eight times greater than the other.
    \item one lower-case letter and even more upper-case letters.
    \item one upper-case letter and starts and ends with a lower-case letter.
    \item one alphanumerical character and contains even more non-alphanumerical characters.
    \item two distinct non-alphanumerical characters, excluding the first and last character. \\
\end{enumerate}
Applying the logical \textit{OR} operator to this enumeration, a binary variable for token $t$ is given by
\begin{align}
    &garbage(t) = 0\text{ if no rule applies,} \nonumber \\
    &garbage(t) = 1\text{ if at least one rule applies.}
\end{align}
Hence, feature $x_i^2$ is extracted from $B_i$ using
\begin{align}
    x_i^2 = f(B_i)[2] = 1-\frac{1}{|B_i^t|} \sum_{t\in B_i^t}garbage(t).
\end{align}
\subsubsection{Publication Year Consideration}
Through BnL data analysis it emerged that the original OCR quality is to some extent sensitive to the period of publication. This property mainly exists due to changes in the OCR engine used and in the source document quality. A yearly basis has been chosen to discretize time. This seems to be the smallest possible time unit that effectively correlates to changes in OCR quality. 
Thus, there is \\
\begin{align}
    x_i^3 = f(B_i)[3] = year(B_i).
\end{align}
\subsubsection{Language Detection and Independence}
It remains to be addressed that the usefulness of two out of the four features, namely $x_0$ and $x_1$, depends on correct language detection. Unfortunately, multilingual data coupled with variable OCR quality renders this task very challenging. BnL tries to overcome this issue by operating on a smaller (text block) level, rather than processing entire articles or pages (with a higher likelihood of language changes). The $langid$ software (\citet{lui}) is used as a fallback after having run $B_i$ against a selection of stop words for $lb$. Also, blocks  where $\ell(B_i)\notin\{de, fr, lb\}$ are discarded (no prediction), opting for a trade-off with a higher accuracy but a little less volume. \\ \\
Although garbage token detection ($x_3$) is not sensitive to the correct detection, it definitely needs to be robust enough to be language independent within $\{de, fr, lb\}$. Since \citet{kulp} and \citet{taghva} based themselves on English data, it was necessary to review the nine listed rules for the languages in question. After careful consideration, it seemed sufficient to reassess rule $3$ and $4$ by verifying that four consecutive vowels and six consecutive consonants are indeed very rare appearances in all three languages.
\subsubsection{Feature Experimentation}
Features that did not contribute to the classifier performance, but were tested by the library, are listed below: \\
\begin{itemize}
    \item A feature stating the font class (Antiqua/Fraktur), derived from the source image.
    \item A metric encoding the value $|B_i|$. Testing was backed by the hypothesis that smaller blocks (mostly headlines) would generally have a lower $x^0$ value induced by the presence of a higher ratio of named tokens not found in $D^{\ell(B_i)}$.
    \item A property indicating $\ell(B_i)$ through one-hot-encoding for a predefined set of language classes.
\end{itemize}
\subsection{Class Definition}
Before a classification model can be created, every $B_i$ needs to be assigned a quality class $Y_i\!\in\!C$ in $T$. Here, the popular Levenshtein $edit$ distance (\citet{levenshtein}) is used to compute quality measure
\begin{align}
    q(B_i)=1-\frac{min(|B_i|, edit(B_i, G_i))}{|B_i|}. \label{formula:quality}
\end{align}
Applying threshold $\theta$ leads to the class definition of
\begin{align}
    \text{if }q(B_i)\geq\theta:&\quad Y_i\!=\!1\!\in\!C, \nonumber \\
    \text{else}:&\quad Y_i\!=\!0\!\in\!C.
\end{align}
\subsection{Implementation} \label{section:implementation}
After having established the computation of $T$, \textsc{Quality} can be fit using a machine learning algorithm. Two non-linear methods were selected for this purpose: \\
\begin{itemize}
    \item K-nearest-neighbour (\textsc{KNN}), motivated by its rather easy implementation, as a first choice.
    \item A feedforward neural network (\textsc{NN}) in view of potentially training a more complex model with more flexibility in terms of hyperparameters. \\
\end{itemize}
The \textsc{NN} architecture showing the best results features two identical $relu$ activated hidden layers with 16 nodes, each followed by dropout of 0.5. Output layer classification is done through $softmax$. Other hyperparameters include a learning rate of $10^{-4}$ and a batch size of 1.
\subsubsection{Preprocessing}
Data standardization is applied in the \textsc{NN} case, for every $d$ from $1$ to $k$, in a way that
\begin{align}
    x^d = \frac{x^d-\Bar{x^d}}{\sigma},
\end{align}
with $\Bar{x^d}$ representing the mean and $\sigma$ the standard deviation. For \textsc{KNN} to guarantee equal importance among features when computing the distance vectors, the feature value ranges need to be equal. That is why better results are obtained trough min-max normalization, i.e.
\begin{align}
    x^d = \frac{x^d-min(x^d)}{max(x^d)-min(x^d)}.
\end{align}
\subsubsection{Training and Testing}
\textsc{Quality} tries to mostly tackle, although influenced by threshold $\theta$, an imbalanced classification problem, with the negative class outnumbering the positive one. This not only makes evaluation of the classifier less trivial, but creates challenges to train on enough positive data points. \\ \\
To perform data augmentation and to specifically combat the lack of positive examples, two \textsc{NewModel} outputs are generated for every block in the ground truth set. \\
\begin{enumerate}
    \item A \textit{new} \textit{best-effort} version, with \textsc{NewModel} being regularly applied, is included in blocks set $B_{new}$.
    \item A \textit{bad} version, with \textsc{NewModel} purposefully applying a model trained on a different font (generating worse results), is included in blocks set $B_{bad}$. \\
\end{enumerate}
\begin{figure}[H]
\centering
\fbox{\includegraphics[width=0.85\textwidth]{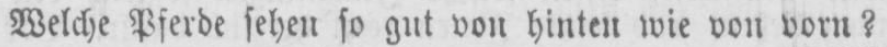}}
\captionsetup{format=hang, margin=0.0\textwidth}
\caption{Very small example block source image.}
\label{fig:pferde}
\end{figure}
Providing examples tied to Figure~\ref{fig:pferde}: \\ \\
\indent{\it "Welche Pferde sehen so gut von hinten wie von vorn?"} $\in B_{new}$, \\
\indent{\it "Welche Pferde sehen so gnt von hinten wie von vorn?"} $\in B_{ori}$, \\
\indent{\it "Belche serde fehen so gut von hinten wie von vorn?"} $\in B_{bad}$.
\\ \\
The sets $B_{new}$ and $B_{bad}$, together with the original OCR output $B_{ori}$, are contained within
\begin{align}
B_{all}=\bigcup\{B_{new}, B_{ori}, B_{bad}\}.
\end{align}
Constant $\alpha$ (directly influenced by the chosen $\theta$) is used to reference the positivity rate, which helps to quantify the imbalance of the problem. To provide an example, $\alpha_{ori}$ denotes the fraction of positive data points within $B_{ori}$. The set $B_{all}$ forms the basis for a training/test set split. A fixed $\beta$ blocks sized test set is first sampled from $B_{all}$ by retaining positivity rate $\alpha_{ori}$, thus creating a realistically imbalanced test scenario. The remaining blocks in $B_{all}$ form the largest possible training set with respect to a perfect $\alpha\!=\!0.5$ rate. In the \textsc{NN} case, $20\%$ of the training set is split-off for validation purposes.
\\ \\
To evaluate \textsc{Quality}, next to the $F_1$ score (harmonic mean of precision and recall), emphasis is put on Cohen's Kappa (\citet{kappa}) metric, which takes class imbalance into account by returning
\begin{align}
    kappa = \frac{p_0-p_e}{1-p_e}. \label{formula:kappa}
\end{align}
In (\ref{formula:kappa}) $p_0$ encodes the accuracy of the test set and $p_e$ is the agreement between the model predictions and the actual class values, as if happening by chance (random class assignment).
\subsubsection{Results}
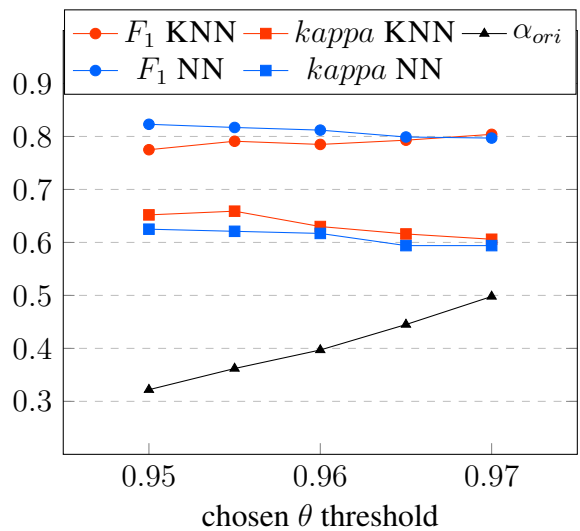
\begin{figure}[H]
\centering
\resizebox{0.48\textwidth}{!}{
\begin{tikzpicture}
\definecolor{c1}{RGB}{255,55,0}
\definecolor{c2}{RGB}{0,100,255}
\definecolor{c3}{RGB}{0,0,0}
\begin{axis} [
    xlabel={chosen $\theta$ threshold},
    xmin=0.945, xmax=0.975,
    ymin=0.2, ymax=1.0,
    xtick={0.95, 0.96, 0.97},
    ytick={0.3, 0.4, 0.5, 0.6, 0.7, 0.8, 0.9},
    legend style={
        at={(0.5,1.05)},
        anchor=north,
        legend columns=3,
    },
    ymajorgrids=true,
    grid style=dashed,
]
\addplot[ % f1 knn
    color = c1,
    mark = *,]
    coordinates {(0.95,0.775)(0.955,0.791)(0.96,0.785)(0.965,0.793)(0.97,0.804)};
\addplot[ % kappa knn
    color = c1,
    mark = square*,]
    coordinates {(0.95,0.652)(0.955,0.659)(0.96,0.630)(0.965,0.616)(0.97,0.606)};
\addplot[ % alpha ori
    color = c3,
    mark = triangle*,]
    coordinates {(0.95,0.322)(0.955,0.362)(0.96,0.397)(0.965,0.445)(0.97,0.498)};
\addplot[ % f1 nn
    color = c2,
    mark = *,]
    coordinates {(0.95,0.823)(0.955,0.817)(0.96,0.812)(0.965,0.799)(0.97,0.797)};
\addplot[ % kappa nn
    color = c2,
    mark = square*,]
    coordinates {(0.95,0.625)(0.955,0.621)(0.96,0.617)(0.965,0.594)(0.97,0.594)};
\legend{$F_1$ \textsc{KNN},$kappa$ \textsc{KNN},$\alpha_{ori}$,$F_1$ \textsc{NN},$kappa$ \textsc{NN}}
\end{axis}
\end{tikzpicture}}
\caption{$F_1$ and $kappa$ scores of \textsc{Quality} given $\alpha_{ori}$ and $\theta$.}
\label{fig:results}
\end{figure}

Results in Figure~\ref{fig:results} are based on $|B_{all}|\!=\!20,166$ and $\beta\!=\!1,000$. Changes in $\alpha_{ori}$ do not seem to affect performance of the classifier significantly, pointing to a rather successfully handled class imbalance. The results can be seen as encouraging, but certainly still leave room for improvement.
\noindent
A hypothesis coupled to experiments conducted with \textsc{Quality}, which potentially explains part of the model errors, states: \\
\begin{align} \label{equation:hypothesis}
    \parbox{32em}{The quality class of smaller sized blocks (e.g. Figure~\ref{fig:pferde}) is considerably harder to determine. This is driven by the reduced amount of data for feature extraction.}
\end{align}
\\
Therefore, the next section's results will make use of this observation by applying a weighted metric.
\section{Enhancement Prediction} \label{section:enhance}
While \textsc{Quality} incorporates a promising start to target an OCR rerun, it does involve a fundamental problem. More specifically for the BnL use case the downside of \textsc{Quality} lies in the lack of enhancement prediction, considering \textsc{NewModel}. Classifying a block as insufficient does not imply that reprocessing changes the class, or even improves the quality at all. Moreover, a binary classifier is prone to provide limited feedback in terms of quality improvement insights. Class conversions alone are not sufficient to obtain a good estimate on the overall improvement of the data.
\subsection{Regression Definition}
Based on this observation, a regression model is leveraged to compute enhancement predictions based on $X_i$. An adequate model naturally needs to output an estimate expressed in the same unit of measure as $q$, as defined in (\ref{formula:quality}), entailing that
\begin{align}
    \textsc{enhance}: B_i \rightarrow [-1,1].
\end{align}
To implement the regression model, $T$ requires one modification. While $X_i$ is left untouched, $Y_i$ is replaced by a continuous variable. Therefore let $i$ and $j$, with $i\!\neq\!j$, denote indices in $B_{all}$. Based on this, all block pairs are enumerated such that $i$ and $j$ reference the same source image and it also holds that
\begin{align}
    &B_i\in B_{ori}\text{ and }B_j\in B_{new}.
\end{align}
Using (\ref{formula:quality}), $Y_i$ is computed in a way that
\begin{align}
    Y_i = q(B_j)-q(B_i)
\end{align}
now encodes the \textit{potential} of the application of \textsc{NewModel}, representing an information that is more valuable to the library while envisioning an OCR rerun.
\subsection{Results}
The machine learning algorithm with the best result is a regression version of \textsc{KNN}, returning the weighted (based on $|B_i|$) mean of all $K$ neighbours. Applied on $T$, \textsc{KNN} outperforms other implementations, such as the same \textsc{NN} architecture (Subsection~\ref{section:implementation}) with adjusted output layer and activation functions, or linear and logistic regressions.
\subsubsection{Performance}
To evaluate \textsc{Enhance}, the mean average error (MAE) measure by \citet{willmott} is used. This provides the ability of interpreting the model performance in the unit of measure $q$. The selected testing method of leave-one-out cross-validation was motivated by the fact that $|T|\!=\!6723$ is relatively small, making it computationally feasible to obtain a robust test result.
\\ \\
Considering $K\!=\!43$ neighbours, $\text{MAE}\!=\!0.034$ is achieved. This can be interpreted as a promising result, given that the test set features a high variance, more precisely a standard deviation of $0.14$. Another reassuring aspect is that the model is only slightly too optimistic, by predicting $0.0029$ too high on average. Overall, no fundamental bias can be observed. 
\\ \\
As stated in (\ref{equation:hypothesis}), predicting on smaller blocks seems to be harder. This hypothesis can be reinforced by evaluating on an adaptation of MAE (here denoted as MWAE), which weights the loss (absolute difference between actual/predicted enhancement) of $B_i$ by $|B_i|$. Since the size of the block obviously directly correlates with the amount of text that is enhanced (or degraded), one can argue that MWAE even represents a fairer evaluation of \textsc{Enhance}. After all, a clear regression performance improvement comes with $\text{MWAE}\!=\!0.024$ for $K\!=\!31$.
\subsubsection{Analysis}
\begin{figure}[H]
\centering
\resizebox{0.5\textwidth}{!}{
\begin{tikzpicture}
\definecolor{c1}{RGB}{255,55,0}
\definecolor{c2}{RGB}{0,100,255}
\definecolor{c3}{RGB}{0,0,0}
\begin{axis} [
    scaled ticks = false,
    tick label style={/pgf/number format/fixed},
    xlabel={chosen $\theta$ threshold},
    ylabel={ratio},
    xmin=-0.11, xmax=0.19,
    ymin=-0.1, ymax=1.2,
    xtick={-0.05,0.0,0.05,0.1,0.15},
    ytick={0.0, 0.2, 0.4, 0.6, 0.8, 1.0},
    legend style={
        at={(0.5,0.98)},
        anchor=north,
        legend columns=3,
    },
    ymajorgrids=true,
    grid style=dashed,
]
\addplot[ % epsilon_r
    color = c1,
    mark = *,]
    coordinates {(-0.06,0.07417793244701087)(-0.04,0.07417793244701087)(-0.02,0.0741544801671887)(0.0,0.07199817332798274)(0.02,0.024519032827980132)(0.04,0.014867442502827302)(0.06,0.008364972196019366)(0.08,0.006129513912413552)(0.1,0.0038008979617363027)(0.12,0.003696991333079701)(0.14,0.0034908067063097055)(0.16,0.0034908067063097055)};
\addplot[ % epsilon_i
    color = c2,
    mark = square*,]
    coordinates {(-0.06,0.0)(-0.04,0.0)(-0.02,0.0001117240552639945)(0.0,0.020421072655162888)(0.02,0.3852743786448752)(0.04,0.6232863549424377)(0.06,0.7349068293038842)(0.08,0.8065923055675712)(0.1,0.8279951167141792)(0.12,0.8362056947347026)(0.14,0.8400639204915598)(0.16,0.845489866009308)};
\addplot[ % c
    color = c3,
    mark = triangle*,]
    coordinates {(-0.06,1.0)(-0.04,1.0)(-0.02,0.9998312738757238)(0.0,0.9764343674920132)(0.02,0.5561307516716264)(0.04,0.30740401503030557)(0.06,0.18895176126621466)(0.08,0.11495884124891208)(0.1,0.0911919100057849)(0.12,0.08286048759895559)(0.14,0.07877751082713585)(0.16,0.07335156530938769)};
\legend{$\epsilon_r$,$\epsilon_i$,$c$}
\end{axis}
\end{tikzpicture}}
\caption{Ratios $\epsilon_r$, $\epsilon_i$ and $c$ for given values of $\theta$.}
\label{fig:results2}
\end{figure}
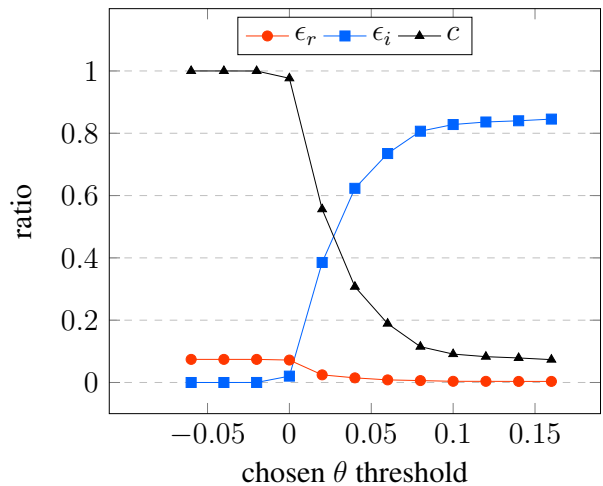

Reprocessing candidate selection based on \textsc{Enhance} requires a cut-off value, here again denoted as $\theta$. Using the policy that every $B_i$, where it holds that
\begin{align}
    \textsc{Enhance}(B_i)\geq\theta,
\end{align}
is selected as a candidate (non-candidate otherwise), three ratios with respect to the total number of blocks, remain of particular importance: \\
\begin{itemize}
    \item The ratio of candidates featuring a strict reduction in $q$, denoted as $\epsilon_r$.
    \item The ratio of non-candidates featuring a strict increase in $q$, denoted as $\epsilon_i$.
    \item The ratio of candidates, denoted as $c$. \\
\end{itemize}
The three ratios (calculated using weighting based on $|B_i|$) are depicted in Figure~\ref{fig:results2} for select values, such that $-0.06\leq\theta\leq0.16$. The graph shows a strong accuracy of \textsc{NewModel} itself (rather low and flat slope of $\epsilon_r$) and ratio $\epsilon_i$ properly adjusting to changes in $\theta$. \\
\noindent Figure~\ref{fig:results2} shows that $\theta$ values, satisfying $0\leq\theta\leq 0.05$, are most suitable for the application of \textsc{Enhance}, considering the BnL data and \textsc{NewModel}.
\\ \\
A Python implementation and data model of \textsc{Enhance}, being part of the source code of the entire OCR project, can be publicly \href{https://github.com/natliblux/nautilusocr}{accessed}\footnote{URL: https://github.com/natliblux/nautilusocr}.
\section{Conclusion}
This article commenced by enumerating three reasons to motivate the requirement of automatic quality assessment. To live up to those needs, feature extraction was discussed by looking at different ideas coming from the literature. Machine learning was applied to build the \textsc{Quality} classifier, designed for block level OCR candidate selection. Finally, this approach was extended through \textsc{Enhance} by considering the potential of \textsc{NewModel}.
\\ \\
The three motivations in (\ref{enum:goals}) can now be re-evaluated as follows: \\
\begin{enumerate}
    \item At the time of writing, BnL already made use of \textsc{Quality} to save processing time. Using $\theta\!=\!0.95$, leading to an appropriate balance in terms of target quality and reprocessing volume, a first experiment was conducted.
    Important to note is the 566,000 newspaper pages (102,000 issues) were processed in merely 15 days. This was enabled by the processing time of \textsc{Quality}, which generally stays below $5\%$ of the time needed for the application of \textsc{NewModel} itself. \\
    \item Next, without statistical insights, the OCR rerun is comparable to a black box. It seems rather unfortunate for new artificial intelligence projects to enable better access to historical data, if those initiatives can not be advertised with concrete numbers. A first BnL application of \textsc{Quality} showed a class change for 70\% of the candidate text lines. Additionally, deep diving into the feature values revealed positive average increments for $x^0$, $x^1$ and $x^2$. However, since this seems insufficient for a very clear picture, \textsc{Enhance} has been developed, expressing its predictions in the most comprehensible unit of measure, being $q$. \\
    \item Lastly, \textsc{Enhance} reinforces the reduction of risks. The selection of candidates for reprocessing based on \textsc{Quality} is exposed to the risk of a poorly performing \textsc{NewModel}. This problem is solved by \textsc{Enhance}, which can be applied using any cut-off threshold, depending on the amount of desired risk. \\
\end{enumerate}
Altogether, \textsc{Enhance} will represent a very helpful addition to the newly developed OCR pipeline of the library and will serve as the basis for future reprocessing candidate selection processes.
\\ \\
The work described in this article has shown that estimating text quality and its potential to improve is a rather difficult task in itself, especially when computational efficiency without source image processing is desired. This is joined by the hurdles of language recognition, the availability of dictionaries covering historical language changes and the challenges involving smaller blocks. Nevertheless, a concrete, applicable and working solution has been proposed. That is why this article was redacted with the intention to share those findings with other cultural institutions with similar requirements.
\newpage
\bibliographystyle{plainnat}
\bibliography{references}
\end{document}